\documentclass[conference]{IEEEtran}
\usepackage[american]{babel}
\usepackage{cite}
\usepackage{amsmath,amssymb,amsfonts}
\usepackage{algorithmic}
\usepackage{graphicx}
\usepackage{textcomp}
\usepackage{xcolor,url}
\usepackage[ruled,vlined,linesnumbered]{algorithm2e}

\newcommand{\cholesky}{\textsc{Cholesky}\xspace}
\newcommand{\potrf}{\texttt{POTRF}\xspace}
\newcommand{\syrk}{\texttt{SYRK}\xspace}
\newcommand{\trsm}{\texttt{TRSM}\xspace}
\newcommand{\gemm}{\texttt{GEMM}\xspace}
\definecolor{potrfc}{HTML}{0191B4}
\definecolor{trsmc}{HTML}{F8D90F}
\definecolor{syrkc}{HTML}{D3DD18}
\definecolor{gemmc}{HTML}{FE7A15}

\usepackage[colorinlistoftodos]{todonotes}

\def\BibTeX{{\rm B\kern-.05em{\sc i\kern-.025em b}\kern-.08em
    T\kern-.1667em\lower.7ex\hbox{E}\kern-.125emX}}

\usepackage{hyperref}

\begin{document}

\title{Geometric deep reinforcement learning \\for dynamic DAG scheduling}

\author{\IEEEauthorblockN{Nathan Grinsztajn}
\IEEEauthorblockA{\textit{Univ. Lille, CNRS, Inria} \\
\textit{UMR 9189 CRIStAL}\\
Lille, France \\
\url{nathan.grinsztajn@inria.fr}}
\and
\IEEEauthorblockN{Olivier Beaumont}
\IEEEauthorblockA{\textit{Hiepacs team} \\
\textit{Inria Bordeaux}\\
Bordeaux, France \\
\url{olivier.beaumont@inria.fr}}
\and
\IEEEauthorblockN{Emmanuel Jeannot}
\IEEEauthorblockA{\textit{TADaaM team} \\
\textit{Inria Bordeaux}\\
Bordeaux, France \\
\url{emmanuel.jeannot@inria.fr}}
\and
\IEEEauthorblockN{Philippe Preux}
\IEEEauthorblockA{\textit{Univ. Lille, CNRS, Inria} \\
\textit{UMR 9189 CRIStAL}\\
Lille, France \\
\url{philippe.preux@inria.fr}}
}

\IEEEoverridecommandlockouts
\maketitle
\IEEEpubidadjcol

\begin{abstract}



  In practice, it is quite common to face combinatorial optimization problems which contain uncertainty along with non determinism and dynamicity. These three properties call for appropriate algorithms; reinforcement learning (RL) is dealing with them in a very natural way. Today, despite some efforts, most real-life combinatorial optimization problems remain out of the reach of reinforcement learning algorithms. 
  
  In this paper, we propose a reinforcement learning approach to solve a realistic scheduling problem, and apply it to an algorithm commonly executed in the high performance computing community, the \cholesky factorization. On the contrary to static scheduling, where tasks are assigned to processors in a predetermined ordering before the beginning of the parallel execution, our method is dynamic: task allocations and their execution ordering are decided at runtime, based on the system state and unexpected events, which allows much more flexibility. To do so, our algorithm uses graph neural networks in combination with an actor critic algorithm (A2C) to build an adaptive representation of the problem on the fly.
  
  We show that this approach is competitive with state-of-the-art heuristics used in high performance computing runtime systems. Moreover, our algorithm does not require an explicit model of the environment, but we demonstrate that extra knowledge can easily be incorporated and improves the performance. We also exhibit key properties provided by this RL approach, and study its transfer abilities to other instances.
  
\end{abstract}

\begin{IEEEkeywords}
  Reinforcement learning, scheduling, task graph, DAG, high performance computing, combinatorial optimization.
\end{IEEEkeywords}

\section{Introduction}

Combinatorial Optimization Problems (COP) constitute an important family of fundamental problems: path finding (\textit{i.e.} traveling salesman problem, vehicle routing problem), stable marriage problem, graph coloring, task scheduling, and many others.
 There are various algorithmic approaches, ranging from (provably) exact methods (\textit{e.g.} based on tree search, linear programming, \textit{etc.}) to non (provably) exact/approximate methods (heuristics and meta-heuristics). Those methods are able to solve large scale COPs, but they require a careful investigation of the problem.
On the other hand, real world applications bring another set of challenges: inherent uncertainty in the definition of the problem and randomness in the process dynamics.
For instance, considering a task scheduling problem, tasks duration and communications delays between tasks are uncertain, and even the very set of tasks to be scheduled may not be known in advance like in an operating system. If we want to tackle real world applications, considering COPs with uncertainty in the settings and the dynamics is necessary.

Reinforcement learning (RL) is designed to deal with sequential decision making under uncertainty \cite{sutton1998introduction}. RL algorithms are able to adapt to their environment: in a changing environment, they adapt their behavior to fit the change. This property opens the door to tackle COPs which contain uncertainty, COPs which are not completely defined when their resolution is initiated. Another potential benefit is the ability to generalize to unseen settings, a necessary step toward real world applications.




In this paper, we will investigate the potential of RL on a real application, the dynamic scheduling of a set of tasks on a distributed computing system.
Modern computer systems contain a variety of resources, interconnected in order to support parallel and distributed computationally intensive applications. Efficiently executing parallel applications on such systems is critical in many scientific domains.
In a high performance computing (HPC) environment, it is very common for an application to be split into several sub-tasks which  may be executed in parallel. There usually are some dependencies between those tasks as the results provided by some may be  necessary to start others. This structure may naturally be modeled using a Directed Acyclic Graph (DAG): nodes of the DAG are sub-tasks, and directed edges represent dependencies. Task-based runtime systems 
internally represent the application as a DAG to execute it on a parallel machine. In this case, one of the main duty of such runtime systems is to schedule the different tasks of the DAG onto the available computing resources. The DAG scheduling problem consists in finding the best way of assigning tasks to processing units, so that the task dependencies are respected and the total duration of execution (the makespan) is minimal.


Scheduling a DAG on a set of resources is a combinatorial problem, known to be NP hard in the strongest sense \cite{NPhard}. It is a sequential, stochastic and dynamic problem. Stochasticity intervenes at two levels: on the one hand, the exact computation times of tasks and transfer times of data are unknown, although we can have good prior estimates. On the other hand, the whole DAG is not necessarily known from the start. This requires that the scheduling algorithm implemented in the runtime system is \emph{dynamic}.
Moreover, in order to be generic, it is important that the proposed solution is able to schedule unknown graphs onto any number of computing resources. 

In this paper, we suppose that different types of tasks can have different durations, but we assume that communications can be neglected (either because the target platform is a shared-memory system or because communications and computations can be overlapped). This allows to simplify the problem formulation while keeping its NP hardness. 
Our contributions are as follows: 

\begin{enumerate}
\item We formalize the dynamic DAG scheduling problem as a Markov Decision Process (MDP).
\item We introduce a practical deep RL algorithm able to build a graph representation sequentially and on the fly.
\item In a set of experiments, we demonstrate that our RL approach obtains schedules competitive with state-of-the-art heuristics, even when no explicit knowledge of the environment is available at the cost of more computation time.
\item We further show that our RL approach is able to generalize, by scheduling yet unseen instances.
\end{enumerate}

\section{Related Works}

Among COPs, task scheduling has attracted a lot of research and presents a rich taxonomy~\cite{leung2004handbook}.
Roughly, it consists in assigning a set of tasks (whose interdependency is represented by a Directed Acyclic Graph -- DAG--) to a set of resources while managing different constraints (a resource cannot execute several tasks at the same time, a task cannot start before its predecessor(s) dependencies have been completed, etc.). In the literature, several classes of scheduling problems have been studied~\cite{graham1979optimization}. In static problems, the DAG is assumed to be completely known in advance while in dynamic problems, part of the DAG is unveiled as the scheduling algorithm progresses. In the homogeneous setting, all resources are identical while in the heterogeneous case, resources are different and the same task can have different duration depending on the resource it is allocated on. Depending on the input (graph topology, number or type of resources) the problem can be either polynomial or NP-Hard. But even simple cases (\textit{e.g.} tasks with no dependencies and two resources) turn out to be NP-Hard~\cite{garey1979computers}.

Applying reinforcement learning to combinatorial optimization has been studied in several articles
\cite{zd1995,LearningCombinatorialOptimizationAlgorithmsoverGraphs,
  CombinatorialOptimizationwithGraphConvolutionalNetworksandGuidedTreeSearch,
  SolvingNP-HardProblemsonGraphsbyReinforcementLearningwithoutDomainKnowledge,
  LearningHeuristicsoverLargeGraphsviaDeepReinforcementLearning} and compiled in this tour d'horizon \cite{tourdhorizon}.
However, performance of RL algorithms facing combinatorial optimization problems remain very far from what traditional approaches and dedicated heuristics achieve. They are mainly distinguished by the graph representation model, the reinforcement algorithm, and the possible use of additional heuristics to help the agent (\textit{e.g}. graph pruning in \cite{CombinatorialOptimizationwithGraphConvolutionalNetworksandGuidedTreeSearch}). The problems studied are classical NP-hard problems, such as the Traveling Salesman Problem (TSP), Minimum Vertex Cover (MVC), and the Max-Cut problem.

Few works focus on reinforcement learning for task scheduling in computational graphs. Most of them~\cite{resiyrcemanagementwithdeeprl,
  multipleResourcemanagement} consider tasks arriving sequentially and randomly, without a DAG structure. The two papers closest to ours are \cite{pmlr-v70-mirhoseini17a,adaptativeDAGtasksscheduling}. The first one uses a very realistic environment (communication time, storage capacity of the nodes...), but relies on a basic tabular Q-learning technique, which cannot scale to real-life applications, and does not allow generalization.
The second one uses deep reinforcement learning, but does not allow online scheduling. Moreover, these two papers preprocess the DAG in ways which do not allow the agent to use its whole structure.

Recently, \cite{paliwal_reinforced_2020} used reinforcement learning to guide a genetic algorithm. However, the scheduling is static, and the environment necessarily deterministic, which are two important limitations with regards to practical constraints.

In~\cite{mao2016resource}, the authors present a reinforcement learning approach to map jobs onto a parallel machine. In contrast to our problem, there are no dependencies between jobs and the goal is to minimize job slowdown.  

In~\cite{pmlr-v70-mirhoseini17a}, the authors study the task mapping of Deep Graph Neural Network. Their approach is based on the policy gradient algorithm. \cite{gao2018spotlight} tackles the same problem as~\cite{pmlr-v70-mirhoseini17a} by modeling it as a Markov decision process and using a reinforcement learning approach called proximate policy optimization~\cite{schulman2015trust}. However, both approaches are not suited for transfer learning as the proposed solutions can only improve the mapping of the input problem.

In~\cite{addanki2019placeto}, the authors provide a general approach for mapping task graph using a graph embedding approach called Placeto. In~\cite{zhou2019gdp}, the authors introduce GDP,  which uses the same approach as Placeto but outperforms it in their experiments. These two approaches allow transfer learning for new graphs but not for new machines (the target platform must be the same as the training one). They also provide the mapping of all tasks at the same time, which is not suitable in case the duration of the tasks are imperfectly known.

Overall, there is yet no generic RL approach that deals with dynamic task graphs and enable generic transfer for both new graphs and new machines. 

\section{Task Modeling}

In this paper, we focus on the \cholesky factorization problem.
The \cholesky factorization is a very common linear algebra routine
along with QR and LU~\cite{choi1996design,buttari2009class}. The
tiled version is used in several task-based runtime systems such as
StarPU~\cite{augonnet2011starpu}
or PARSEC~\cite{bosilca2013parsec}. These runtime systems are in charge of scheduling
the tasks onto homogeneous or heterogeneous platforms based on the description of the application by means of a DAG (as
depicted in Fig.~\ref{fig:chol_5_dag}). Hence, being able to efficiently
schedule the \cholesky DAG is of utmost importance. It is indeed characteristic of many applications in linear algebra and scientific computing, in the sense that \cholesky factorization involves (i) a large number of tasks, (ii) complex but regular dependencies and (iii) a small number of different kernels \footnote{in this paper, a ``kernel'' is a
  basic operation performed on a sub-matrix. The \cholesky
  factorization algorithm is expressed as a combination of 4
  different kernels.}. It is therefore a very good benchmark for scheduling algorithms~\cite{agullo2016static,agullo2010dynamically,jeannot2013symbolic} and designing a scheduling algorithm for the dense tiled \cholesky factorization is paramount, both practically and
theoretically. 

\subsection{Tiled \cholesky Factorization}

In this paper we focus on the task graph induced by the tiled dense \cholesky factorization depicted in Algorithm~\ref{algo:cholesky}


\begin{algorithm}[htb]
{\small
  \For{$k = 0 ... T -1$}{
  $A[k][k] \leftarrow \mbox{\color{potrfc}\bf\potrf}(A[k][k])$\\
  \For{$m = k+1 ... T -1$}{
  $A[m][k] \leftarrow \mbox{\color{trsmc}\bf\trsm}(A[k][k],A[m][k])$ \label{line:trsm}\\ 
  }
  \For{$n = k+1 ... T -1$}{
  $A[n][n] \leftarrow \mbox{\color{syrkc}\bf\syrk}(A[n][k],A[n][n])$ \\
    \For{$m = n+1 ... T -1$}{
      $A[m][n] \leftarrow \mbox{\color{gemmc}\bf\gemm}(A[m][k],A[n][k],A[m][n])$\\ 
    }
  }
}
}
\caption{Tiled version of the \cholesky factorization.}\label{algo:cholesky}
\end{algorithm}

For a given symmetric positive definite matrix $A$,
the \cholesky algorithm computes a lower triangular matrix $L$ such that
$A=LL^T$. In the tiled version, the matrix is decomposed into $T\times T$
square tiles. Each tile is hence a sub-matrix of the original
matrix. We denote $A[i][j]$ the tile corresponding to row $i$ and column $j$: the
reader should be careful that this $A[i][j]$ is a $N\times{}N$
sub-matrix of the original matrix, made of the elements of rows and
columns (\textit{i.e.} spanning rows from $N\times{}i$ to $N\times (i+1)-1$).
$N$ 
is usually several hundreds (typically $380\times 380$ for standard CPUs and $960 \times 960$ for GPUs). At each
step $k$, the algorithm performs a \cholesky factorization of the tile
$A[k][k]$ located on the diagonal 
(with the \potrf kernel). Then it updates all the tiles below it
($A[k][k+1:T-1]$)  using a triangular solve (\trsm kernel). The trailing sub-matrix is updated
using  the \syrk kernel for tiles on the diagonal and matrix multiply (\gemm
kernel) for the remaining tiles (of the lower triangular part).

\begin{figure}[htbp]
  \centerline{\includegraphics[width=0.95\linewidth]{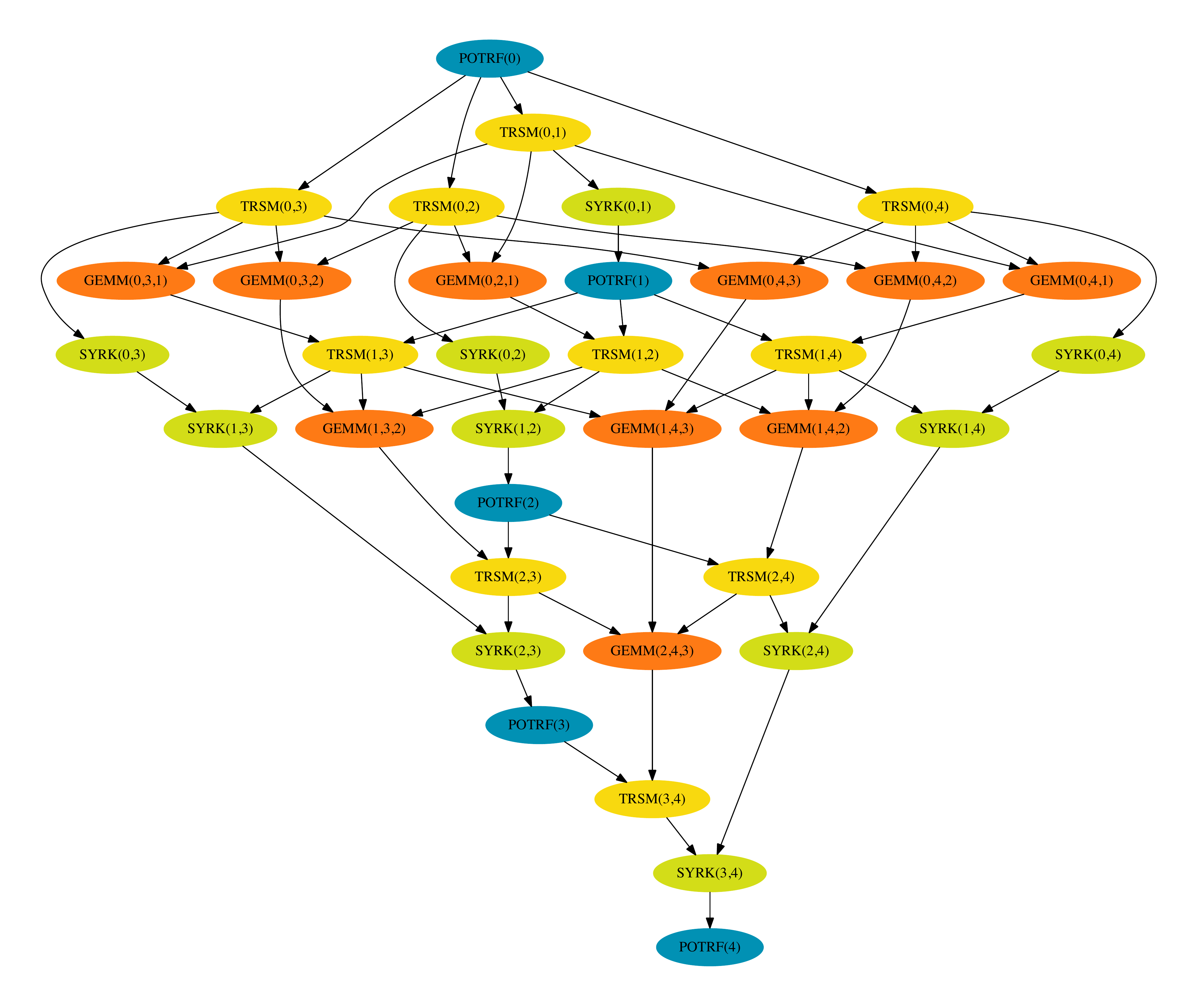}}
\caption{DAG of the \cholesky Factorization for $T=5$. The indices of \potrf\unskip$(k)$,
  \trsm\unskip$(k,m)$, \syrk\unskip$(k,n)$ and \gemm\unskip$(k,n,m)$ correspond to the loop
indices of Algorithm~\ref{algo:cholesky}.}
\label{fig:chol_5_dag}
\end{figure}

Each kernel \potrf, \trsm, \syrk and \gemm is therefore executed
several times during the \cholesky factorization by different {\em
  task instances}. Each task instance requires input data produced by
other tasks. This leads to a DAG in which nodes are task instances and
directed edges are (data) dependencies between two task
instances. For example, for $k=0$ the \potrf kernel updates tile
$A[0][0]$. This tile is then used by all the \trsm kernel to update
tiles $A[1:T-1][k]$. Hence, in the graph there are $T-1$ edges between
the task that instantiate the \potrf kernel and $T-1$ tasks that each
instantiate a \trsm kernel. 
Fig.~\ref{fig:chol_5_dag} illustrates graphically the \cholesky
DAG for $T=5$ (\textit{i.e.} 5 by 5 tiles).

The advantage of the tiled version of the \cholesky factorization is
three-fold. First, by working on tiles, the computation of the kernels is
very fast and optimized using BLAS (Basic Linear Algebra Subroutine) kernels. Second, tiles
enable to deal with large size problems with relatively small
DAGs. Third, the tiled version expresses a lot of parallelism which
facilitates its execution on modern parallel system.  

\subsection{Problem Definition}
\label{sec.pbdef}

Our scheduling problem can be formalized as follows. For a given value
of $T$, the \cholesky factorization is represented by a DAG, $G = (V, E)$, where the set of vertices $V$
corresponds to the set of $n$ tasks in the application, and $E$ is the set of
directed edges between tasks, expressing dependencies between the result of the execution of a task, and its use by a subsequent task. One vertex corresponds to the beginning of the factorization process, while an other vertex corresponds to its completion. Each task can be either one among the four types (of kernels) that
determine its duration. The duration of a task can be deterministic or stochastic. For simplicity, we will consider them deterministic,
although our optimization method works the same way for stochastic
durations. We denote by $d(v)$ the duration of the task $v$, and by $W = \sum_{v \in V} d(v) $ the total work needed to complete the
instance. %
In what follows, we assume that the
communication cost due to dependencies is zero. This is justified
by the two following use-cases.  
First, if the target machine is a multi-core system with shared memory,
each core can execute one kernel at a time and the whole matrix is
stored in a memory shared by all the cores. Hence, communication
between task is done through memory access and such accesses are
negligible compared to kernel computation time.
Second, if the target machine consists of GPUs, we can notice that for
a tile of 
size $N$, the amount of data to transfer is of the order of $N^2$
while the complexity of the different kernels is of the order of
$N^3$, so that one generally chooses a tile size $N$ large enough to
overlap computations and communications, while allowing an efficient
use of computation resources (cache,...).

\label{sec:cp}
The \emph{makespan} is defined as the completion time of the last task to be executed. 
Two notions are of importance in what follows. The critical path $CP$
of a DAG is the longest distance between the start node and the end node,
including all the tasks and their duration. The total work ratio is
equal to $\frac{W}{P}$ (where $p$ is the number of computing
resources of the parallel machine). $CP$ and $\frac{W}{P}$ are both lower bounds of the optimal makespan. Our objective is to minimize the makespan.

\subsection{Reinforcement Learning Formulation}

We model the problem as a Markov Decision Process (MDP). Given a state $s$, an agent chooses an action $a$ to maximize a reward function $R$. A MDP can thus be broken into four parts:
\begin{itemize}
    \item a set of states $S$,
    \item a set of actions, $A$,
    \item $P_{a}(s,s')=\mathbb{P}(s_{t+1}=s'\mid s_{t}=s,a_{t}=a) $ is the probability of the transition from state $s$ to state $s'$ under the action $a$.
    \item $R_{a}(s,s')$ is the immediate reward after the transition from $s$ to $s'$ with the action $a$.
\end{itemize}

Given $S$, $A$, $P$, and $R$, resolving the MDP consists in finding a policy that optimizes a certain objective function.
We now define $S$, $A$, $R$, and the objective function used to model the DAG scheduling problem.

\subsubsection*{State Space}

The goal is to give the agent as much information about the task DAG as necessary. Computing the optimal solution requires the whole graph, and therefore the ``state'' should embed the whole graph. However, the whole DAG being potentially arbitrary large and too cumbersome to be handled in practice, we consider an approximate representation. Hence, we restrict the information represented in a state to information about running tasks and available tasks, along with their descendants: 
``running'' tasks are those currently executed, ``available'' those that may be executed but can not because of a lack of computing resources, ``descendants'' are all the tasks that have to wait for running and available tasks to be completed to be run, because they depend on their results. The depth of descendants being considered is left as a parameter in our algorithm; it is denoted by the window $w$. This choice of $w$ corresponds to a trade-off between computational time and accuracy.


Each node is represented by a set of raw features: these features are expected to encode and summarize the DAG information at the node level. The representation $X_i$ of node $i$ can be written:
\[ X_i = \lbrack succ_i, pred_i, type_i, avail_i, run_i, cp_i \rbrack \]
where $succ_i$ is the number of successor nodes of $i$, $pred_i$ is the number of predecessor nodes of $i$, $type_i$ is the type of the task encoded as an one-hot vector, $avail_i$ is a binary variable indicating if the task is available, $run_i$ is a binary variable indicating if the task is currently running, and $cp_i$ contains the portion of critical path ahead of the task.
CNNs are commonly used to process images in deep RL. CNNs exchange information between neighboring pixels, resulting in smoothing pixels. Likewise, we use a particular convolution tailored for graphs to exchange information between neighboring nodes of a graph, namely graph convolution networks (GCN)~\cite{GCN} which have been shown to perform well on several graph-related task benchmarks. 
Stacking such (graph) convolution layers mixes the information of nodes located always further from each other.
At the output of the last convolution layer, we obtain an internal representation of the input of the network; an embedding has been computed for each node.
We refer the interested reader to the description of GCN~\cite{GCN} for further details.

To summarize, from the input state of the network which is a sub-DAG, an internal state representation is computed by a set of stacked graph convolution layers.

\subsubsection*{Action Space}

An action consists in selecting an available task or in doing nothing (pass). If a task $t$ is selected, it is immediately scheduled on a ready-for-use processor. If all processors are ready, it is not possible to pass.

\subsubsection*{Reward and objective function}

There is no reward (i.e $r_t=0$) except when the final state is reached, or equivalently when the whole input DAG has been computed. Indeed, there is no relevant information available before the whole DAG has been scheduled that could be used as an immediate reward.
The reward uses the final makespan given by the whole scheduling trajectory, normalized by a baseline duration. We use a heuristic (the ASAP algorithm, detailed in Section~\ref{sec.refasap}) to get a baseline duration in this work.

Thus, the reward can be written as: 
\[ R(makespan) = \frac{makespan_{ASAP} - makespan}{makespan_{ASAP}} \]

The lower the makespan the better; therefore the reward becomes positive as soon as the learned policy becomes more efficient than ASAP.

\section{Algorithm}
\subsection{Actor Critic}
\label{A2Cformalism}

We train an agent to schedule tasks using synchronous actor critic (A2C, \cite{A2C}). A2C is a policy gradient method which aims at maximizing the objective function directly by gradient ascent: 
\begin{equation*}
    J(\theta) =  \mathbb{E}_{\tau \sim \pi_{\theta}} \left( \sum _t \gamma^t r(s_t, a_t) \right)
    \end{equation*}
where $\tau$ is a trajectory sampled according to the policy $\pi$ parametrized by $\theta$, and $\gamma$ a discount factor.

To compute a single update, we first run the current policy up to $t_{max}$ steps or until a terminal state is reached.

A2C uses a policy network (the actor) $\pi_\theta(a_t \mid s_t)$, parameterized by $\theta$, which computes a distribution of probability over the actions, and a value network (the critic) which estimates the value of a state $V_{\theta_v}(s_t \mid \theta_v)$, and is used to lower the variance in the computation of the advantage function.

The policy update takes the form $\nabla_{\theta} \log \pi_\theta(a_t \mid s_t) A(s_t, a_t, \theta, \theta_v)$, where the advantage function $A(s_t, a_t, \theta, \theta_v)$ can be written $\sum_{i=0}^{k-1} \gamma^i r(s_{t+i}, a_{t+i}) + V_{\theta_v}(s_{t+k}) - V_{\theta_v}(s_{t})$. $k$ is either the time when a terminal state is reached, or $t_{max}$. In the first case, $V_{\theta_v}(s_{t+k}) = 0$.
The critic is simply updated in order to minimize the mean square error between the predicted value $V_{\theta_v}(s_{t})$ and the real return 
$\sum_{i=0}^{k-1} \gamma^i r(s_{t+i}, a_{t+i}) + V_{\theta_v}(s_{t+k})$.

Adding the entropy of the policy in the objective function has been shown to improve exploration \cite{A2Centropy}. We therefore add the term $\beta \nabla_\theta H(\pi_\theta(s_t))$ to the actor gradient, where $H$ is the entropy and $\beta$ a hyper-parameter which controls the influence of the entropy regularization.

In practice, many parameters of the actor and the critic are shared, as we will detail in the next section.

\subsection{Architecture}

\begin{figure}[htbp]
  \centerline{\includegraphics[width=0.95\linewidth]{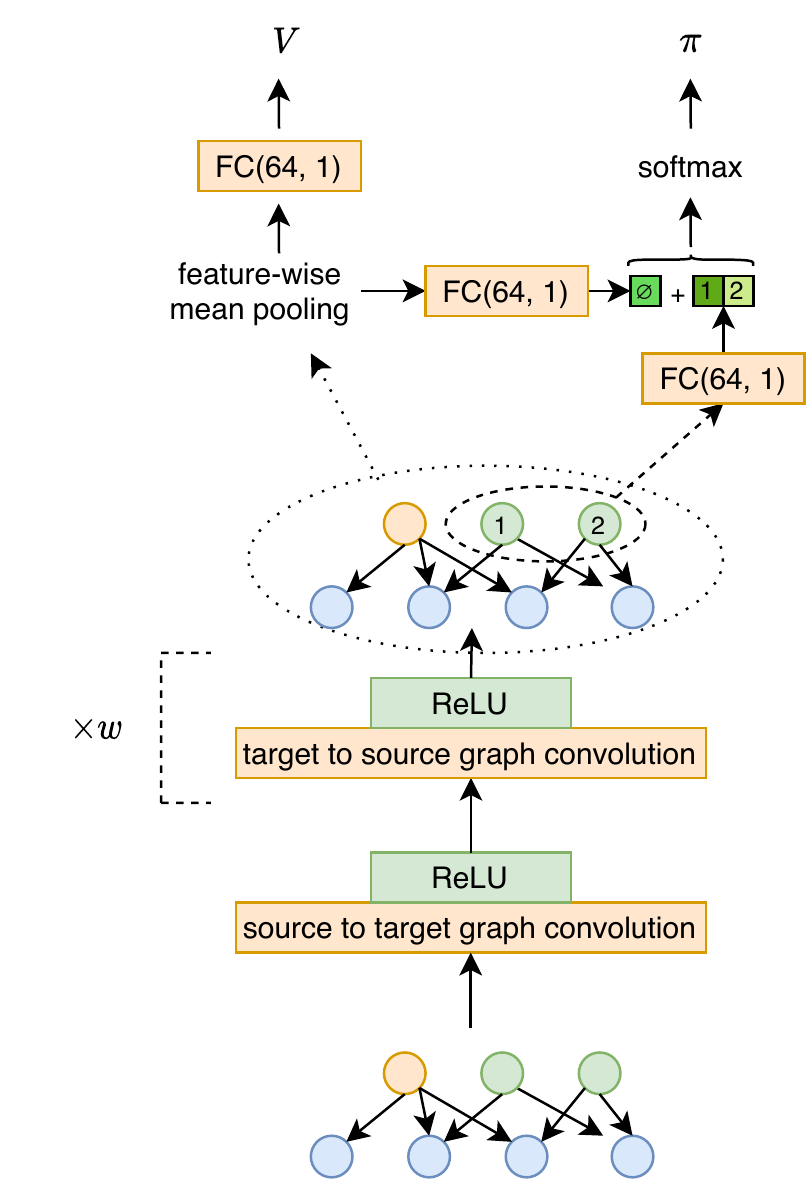}}
  \caption{Overview of the agent architecture. At the bottom, a sub-DAG is input into a stack of $1+w$ graph convolution layers. An internal representation of the sub-DAG is output in which node information has been mixed. For this current state, an estimate of the value $V$ and an action to perform (either do nothing ($\emptyset$) or begin the execution of one of the available tasks - in green, either 1 or 2) are generated. FC(64, 1) is a fully connected layer with an input size of 64 and an output size of 1.
  }
  \label{fig:narch}
\end{figure}

The network architecture is kept as simple as possible in order to minimize the scheduling computation overhead (see Fig.~\ref{fig:narch}).

A sub-DAG is input in the neural network (bottom of Fig.~\ref{fig:narch}) and goes through a series of graph convolution layers.
The number of layers is a parameter of the algorithm; it should be related to the size of the window $w$: 
at least $w$ layers are required to let the necessary information flow among the nodes of the sub-DAG. Empirically, we found that using exactly $w$ iterations is enough. 
Between these layers, we use ReLU functions as non-linear activations.

As already mentioned above, these stacked convolution layers produce an internal representation of the input sub-DAG. This representation is used to produce an estimation of the value of the current state (the input sub-DAG), and an action to perform. A set of 3 1-layer fully-connected networks (FC) produce these two results.

This architecture being quite complex and impossible to present with all the details needed to re-implement it, we provide our own implementation. Please refer to the experiment section.



\subsection{Scheduling}

The scheduling process is done iteratively by placing the available task chosen by the agent on an available device. Once every device has been assigned a task or the agent has decided to pass, there is no environment-agent interaction until the next event, the moment when one or more tasks are completed, and the corresponding computing units become available. Then the agent can choose a new action, and the process goes on until the whole DAG has been computed.

\section{Experiments}

\subsection{Reference Algorithm} 
\label{sec.refasap}

In this paper, our goal is to analyze the performance of a reinforcement learning based algorithm for the dynamic scheduling of \cholesky factorization tasks. The dynamic nature of factorization is in practice imposed by the difficulty of accurately predicting computational costs and communication durations in an HPC environment in which the various operations unpredictably influence execution times. Since it is not possible to schedule and allocate tasks in advance, in practice, dynamic runtimes rely solely on the description of the machine state and on the tasks already performed, using a task priority system to define which tasks to perform in the event that the number of available resources is less than the number of available tasks. In these dynamic systems, task placement decisions are made a little in advance, taking into account the placement of input data, and this delay is used to transfer task input data if necessary to overlap communication and computation. We have already discussed in Section~\ref{sec.pbdef} how to overlap communications and computations in both the CPU multicore and the GPU cases.

To perform dynamic scheduling, ASAP is a strategy of choice which is the \textit{de facto} standard in most dynamic schedulers. ASAP never leaves a resource inactive if there is an executable task and ASAP chooses among several candidate tasks the one that is the farthest from the end of the computation (the one with the longest critical path). It has been demonstrated in~\cite{europar20} that despite its simplicity, this strategy gives excellent results for \cholesky factorization, especially in the case where execution times are similar to what is observed in practice on GPUs. This is therefore the strategy that we use as a baseline to evaluate the performance of the reinforcement learning based algorithm that we propose: the reader should keep in mind that it is difficult to beat ASAP, or even to perform as well as ASAP. However, ASAP requires the whole DAG: ASAP cannot cope with a dynamic environment in which the DAG is unknown. Our approach does not suffer from this limitation.

We add two other baselines, Random and Greedy. We call Greedy the baseline which prioritizes the tasks which have the largest number of successors. Random consists simply in choosing the task to schedule uniformly among the available tasks. Both baselines are very simple, hence computed very quickly.

For reproducibility purposes, the code used to perform the experiments is available freely on the web at \url{https://github.com/nathangrinsztajn/DAG-scheduling}. 

\subsection{Simulated Model}

In order to be able to iterate rapidly over runs, we do not evaluate the agent performance on a real device but on a simulated environment. We use a different mean duration for each type of task, as shown in table \ref{tab_time}, according to the data gathered in \cite{europar20} for GPU computations.
\begin{table}[htbp]
  \caption{Task durations used in the DAG model.}
\begin{center}
\begin{tabular}{|c|c|c|c|}
\hline
\mbox{\color{potrfc}\bf\potrf} & \mbox{\color{syrkc}\bf\syrk} & \mbox{\color{trsmc}\bf\trsm} & \mbox{\color{gemmc}\bf\gemm} \\
\hline
11 & 2 & 8 & 3 \\
\hline
\end{tabular}
\label{tab_time}
\end{center}
\end{table}

\subsection{Results}

We perform different types of simulations:

\begin{enumerate}
\item performance comparison of our RL approach with several baselines, using \cholesky factorization with different numbers of tiles.
\item a closer look at the performances if we remove the critical path of the node embedding and vary the window parameter $w$.
\item transfer learning: we perform different kinds of transfers:
  \begin{enumerate}
  \item having learned to schedule the DAG of tasks of a \cholesky factorization for a given number of tiles $T$, how does this transfer to other numbers of tiles?
  \item having learned to schedule the DAG of tasks of a \cholesky factorization for a given number of computing devices $p$, how does this transfer to other numbers of computing devices?
  \end{enumerate}
\end{enumerate}

As our approach is not deterministic, we train 10 agents for each configuration with 10 different seeds. The graph neural network policy was developed using PyTorch Geometric package \cite{pytorchgeomtric} and trained on the 8 cores of a CPU (no GPU were used), a run taking approximately 1 hour to complete. In the tables, we provide the results of the best of the 10 agents.

Using the notations introduced in section \ref{A2Cformalism}, we set $\beta = 0.02$, $t_{max} = 40$, and $\gamma = 1$ 
in all our experiments. Each agent is trained for 10,000 steps and only the best version encountered during the process is kept.
Training was done using Adam optimizer, with a learning rate of $0.01$ and $\epsilon = 0.1$. As some parameters of the actor and the critic are shared, we give the actor update more importance by down-weighting the critic learning rate by $1/2$.

Unless specified otherwise, we take $w=1$ for training and testing.

We use 3 different DAG obtained for three different numbers of tiles $T$: 4, 8 and 16. These graphs have very distinct characteristics, as shown in Table \ref{featureswithn}. $\lvert V \rvert$ is the number of nodes in the DAG. Please refer to section \ref{sec:cp} for the definition and meaning of $W$ and critical path.

\begin{table}
\caption{DAG characteristics for several number of tiles $T$ in the \cholesky factorization.}
\begin{center}
\begin{tabular}{|c|c|c|c|}
\hline
$\boldsymbol{T}$ & $\boldsymbol{\lvert V \rvert}$ & $\boldsymbol{W}$ &  \textbf{Critical Path} \\
\hline
4 & 21 & 116 & 74 \\
\hline
8 & 121 & 536 & 158 \\
\hline
16 & 817 & 3056 & 326 \\
\hline
\end{tabular}
\label{featureswithn}
\end{center}
\end{table}

\begin{table}[htbp]
  \caption{Makespan comparison (lower is better). For stochastic baselines, the result is a mean over 10 trajectories, and the standard deviation is given in parentheses.}
\begin{center}
\begin{tabular}{|c|c|c|c|c|c|}
\hline
$\boldsymbol{T}$ & $\boldsymbol{p}$ & \textbf{Agent} & \textbf{ASAP} & \textbf{Greedy} & \textbf{Random}  \\
\hline
4 & 4 & \textbf{74} & \textbf{74} & \textbf{74} & 74.8 (0.87)\\
\hline
8 & 4 & 163 & \textbf{160} & 173 & 196.5 (5.57) \\
\hline
16 &  4 & 792 & \textbf{787} & 814 & 832.9 (6.09) \\
\hline
8 & 2 & \textbf{280} & 282  & 286 & 300.2 (5.39) \\
\hline
8 & 6 & \textbf{158} & \textbf{158}  & 174 & 174.2 (3.24)  \\
\hline
\end{tabular}
\label{tab_results1}
\end{center}
\end{table}

\subsubsection{RL \textit{vs}. ASAP: performance comparison}

Table \ref{tab_results1} reports the performance of the baselines and our RL approach for different numbers of tiles $T$ and processors $p$. Performance is measured as the makespan of the scheduling computed by a given algorithm. We can see that our RL approach is very competitive with ASAP, and outperforms consistently the other baselines. Again, the reader should keep in mind that ASAP is a very good and hard to beat heuristic.

\subsubsection{A closer look at $w$ and our initial embedding}

The length of the critical path ahead added in the initial node embeddings gives information about all the tasks remaining at each step, and supposes an accurate model of the DAG and of the sub-task durations. As this information is not always available, we investigate the performance of our algorithm without such a help: the results are given in Table \ref{tab_results2}. We choose to compare those configurations with $T=8, p=4$ as it is one of the most difficult RL settings, according to Table \ref{tab_results1}.

\begin{figure}[htbp]
  \centerline{\includegraphics[width=0.95\linewidth]{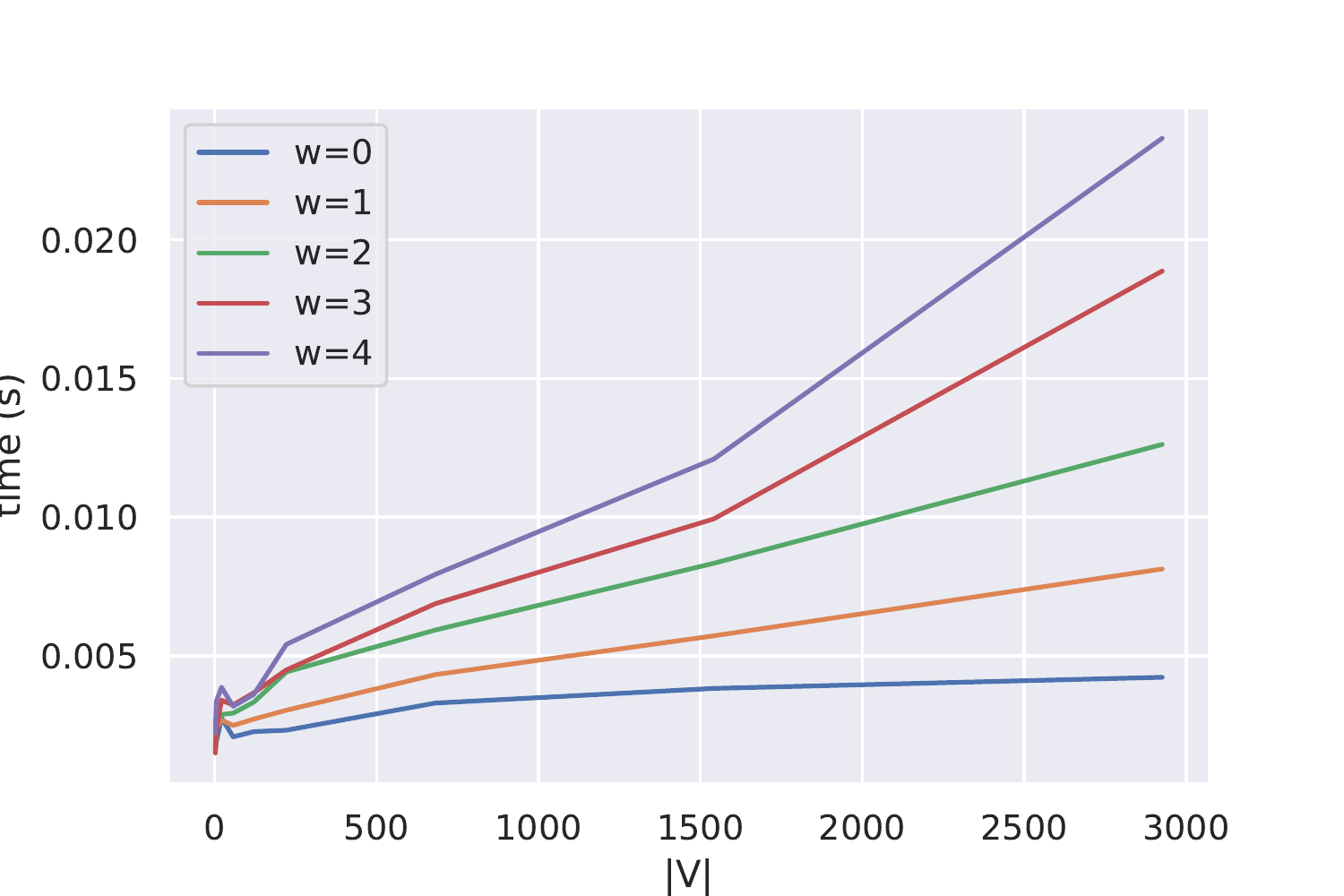}}
  \caption{Mean computation time per action at inference, for several $w$ and several DAG sizes. The computation time per action increases with the number of nodes in the graph and with $w$, as in both cases there are more message-passings to compute during the updates of the node embeddings.}
  \label{fig:time_window}
\end{figure}

We observe that when the critical path is included in the embedding, there is no benefit in enlarging $w$. In fact, $w=0$ is already almost optimal. On the contrary, without this information, we note that enlarging $w$ greatly improves the performances and seems to have a stabilizing effect on the training. When $w=4$, the makespan is almost as good as those of agents trained with the critical path. On the other hand, rising $w$ increases the computation time, as can be seen in Fig.~\ref{fig:time_window}. $w$ should therefore be tuned adequately to ensure a suitable time-performance trade-off for the current case of use. ASAP cannot be used when the critical path is not available.

\begin{table}[htbp]
  \caption{Performance comparison of several settings. The CP column contains $+$ if we included the critical path in the node embeddings, and $-$ otherwise. We run each configuration 10 times and keep for each one the results of best agent. The standard deviation of the makespans of the 5 best agents is written in parentheses.}
\begin{center}
\begin{tabular}{|c|c|c|c|c|}
\hline
$\boldsymbol{T}$ & $\boldsymbol{p}$ & \textbf{CP} & $\boldsymbol{w}$ & \textbf{makespan}  \\
\hline
 &  & $+$ & 0 & 163 (3.28) \\
\cline{3-5}
 &  & $+$ & 1 & 163 (4.54) \\
\cline{3-5}
 &   & $-$ & 0 & 173 (40.13) \\
\cline{3-5}
8 & 4 & $-$ & 1  & 170 (16.53) \\
\cline{3-5}
 &  & $-$ & 2  & 171 (0.89) \\
\cline{3-5}
 &  & $-$ & 3  & 166 (0.83) \\
\cline{3-5}
 &  & $-$ & 4  & 164 (10.58) \\
\hline
\end{tabular}
\label{tab_results2}
\end{center}
\end{table}

\subsubsection{Experiments on transfer learning}

Table \ref{tab_results3} reports makespans achieved with transfer learning: we train our agent on a given number of tiles $T$, and we measure the makespan achieved for two other values. When $T_{\mbox{train}}$ and $T_{\mbox{test}}$ are equal, the results are those reported in table \ref{tab_results1}. We can see that the transfer is very effective: the makespan obtained with this 0-shot transfer is only slightly worse than the one of a dedicated agent, and still not much worse than ASAP.
We conduct the same experiments with the number of computation units $p$. We notice that our agent exhibits good transfer abilities across $p$, beating the Greedy baseline almost in every case .
While training a RL agent takes time, this experiment shows that once trained, the RL agent has the ability to schedule different DAGs, something ASAP cannot do.

\begin{table}[htbp]
  \caption{Transfer learning experiments through $T$ (number of tiles) and $p$ (number of processing units). The RL agent learns on $T_{\mbox{train}}$ and $p_{\mbox{train}}$ and is tested at inference on $T_{\mbox{test}}$ and $p_{\mbox{test}}$.}
\begin{center}
\begin{tabular}{|c|c|c|c|c|}
\hline
$\mathbf{T_{\mbox{test}}}$ & $\mathbf{T_{\mbox{train}}}$ & $\mathbf{p_{\mbox{test}}}$ & $\mathbf{p_{\mbox{train}}}$ & \textbf{makespan} \\
\hline
 & 4  & & & 74 \\
4 & 8 & & & 74 \\
 & 16 & & & 74 \\
\cline{1-2}
\cline{5-5}
 & 4 & & & 215 \\
8 & 8 & 4 & 4 & 163 \\
 & 16 & & & 175 \\
\cline{1-2}
\cline{5-5}
 & 4 & & & 911 \\
16 & 8 & & & 805 \\
 & 16 & & & 792 \\
\hline
 &  &  & 2 & 280 \\
 &  & 2 & 4 & 285 \\
 &  &  & 6 & 296\\
  \cline{3-5}
 &  &  & 2 & 172\\
8 & 8 & 4 & 4 & 163 \\
 &  &  & 6 & 178 \\
  \cline{3-5}
 &  &  & 2 & 158\\
 &  & 6 & 4 & 159 \\
 &  &  & 6 & 158\\
 \hline
\end{tabular}
\label{tab_results3}
\end{center}
\end{table}

\section{Conclusion and Future Work}

In this paper, we have investigated the use of reinforcement learning as a principled approach to solve scheduling problems involving the need of being able to adapt to a dynamic (runtime) environment. Solving scheduling problems is known to be NP-hard, and remains a challenge for RL. In this paper, we focused on dynamic scheduling on homogeneous resources without communication costs. We conducted our experiments on a well-known, heavily used numerical procedure, the \cholesky factorization. Experiments show that we can achieve schedules that are as efficient as those obtained by dedicated heuristics; they also show the benefit of using a RL approach to transfer the policy learned on a certain hardware configuration to another or to transfer the scheduling policy learned on a certain graph of tasks to another.
To the best of our knowledge, our paper is the first to present such an adaptive/RL approach featuring dynamic scheduling and transfer learning, and to study the effect of the node-level information available.







There are numerous directions to go further. As we only considered homogeneous computing resources, examining heterogeneous devices such as CPUs and GPUs would be interesting. Considering other types of tasks, such as LU factorization, which is more complex than \cholesky because of the repeated selection of a pivot, is also a path to investigate. Yet another direction is the integration of our RL approach in a runtime system.



\newpage
\IEEEtriggeratref{17}
\bibliographystyle{plain}
\bibliography{ref}

\begin{thebibliography}{10}

\bibitem{SolvingNP-HardProblemsonGraphsbyReinforcementLearningwithoutDomainKnowledge}
Kenshin Abe, Zijian Xu, Issei Sato, and Masashi Sugiyama.
\newblock Solving {NP}-hard problems on graphs by reinforcement learning
  without domain knowledge.
\newblock {\em CoRR}, abs/1905.11623, 2019.

\bibitem{addanki2019placeto}
Ravichandra Addanki, Shaileshh~Bojja Venkatakrishnan, Shreyan Gupta, Hongzi
  Mao, and Mohammad Alizadeh.
\newblock Placeto: Learning generalizable device placement algorithms for
  distributed machine learning.
\newblock {\em arXiv preprint arXiv:1906.08879}, 2019.

\bibitem{agullo2010dynamically}
Emmanuel Agullo, C{\'e}dric Augonnet, Jack Dongarra, Hatem Ltaief, Raymond
  Namyst, Jean Roman, Samuel Thibault, and Stanimire Tomov.
\newblock {Dynamically scheduled Cholesky factorization on multicore
  architectures with GPU accelerators}.
\newblock In {\em Symposium on Application Accelerators in High Performance
  Computing (SAAHPC)}, 2010.

\bibitem{agullo2016static}
Emmanuel Agullo, Olivier Beaumont, Lionel Eyraud-Dubois, and Suraj Kumar.
\newblock {Are static schedules so bad? A case study on Cholesky
  factorization}.
\newblock In {\em 2016 IEEE International Parallel and Distributed Processing
  Symposium (IPDPS)}, pages 1021--1030. IEEE, 2016.

\bibitem{augonnet2011starpu}
C{\'e}dric Augonnet, Samuel Thibault, Raymond Namyst, and Pierre-Andr{\'e}
  Wacrenier.
\newblock {StarPU: a unified platform for task scheduling on heterogeneous
  multicore architectures}.
\newblock {\em Concurrency and Computation: Practice and Experience},
  23(2):187--198, 2011.

\bibitem{europar20}
Olivier Beaumont, Julien Langou, Willy Quach, and Alena Shilova.
\newblock {A Makespan Lower Bound for the Scheduling of the Tiled Cholesky
  Factorization based on ALAP scheduling}.
\newblock In {\em Proceedings of the 26th International European Conference on
  Parallel and Distributed Computing (EuroPar 2020)}, pages 1--12, 2020.

\bibitem{tourdhorizon}
Yoshua Bengio, Andrea Lodi, and Antoine Prouvost.
\newblock Machine learning for combinatorial optimization: a methodological
  tour d'horizon.
\newblock {\em CoRR}, abs/1811.06128, 2018.

\bibitem{bosilca2013parsec}
George Bosilca, Aurelien Bouteiller, Anthony Danalis, Mathieu Faverge, Thomas
  H{\'e}rault, and Jack~J Dongarra.
\newblock Parsec: Exploiting heterogeneity to enhance scalability.
\newblock {\em Computing in Science \& Engineering}, 15(6):36--45, 2013.

\bibitem{buttari2009class}
Alfredo Buttari, Julien Langou, Jakub Kurzak, and Jack Dongarra.
\newblock A class of parallel tiled linear algebra algorithms for multicore
  architectures.
\newblock {\em Parallel Computing}, 35(1):38--53, 2009.

\bibitem{choi1996design}
Jaeyoung Choi, Jack~J Dongarra, L~Susan Ostrouchov, Antoine~P Petitet, David~W
  Walker, and R~Clint Whaley.
\newblock Design and implementation of the {ScaLAPACK LU, QR, and Cholesky}
  factorization routines.
\newblock {\em Scientific Programming}, 5(3):173--184, 1996.

\bibitem{LearningCombinatorialOptimizationAlgorithmsoverGraphs}
Hanjun Dai, Elias~B. Khalil, Yuyu Zhang, Bistra Dilkina, and Le~Song.
\newblock Learning combinatorial optimization algorithms over graphs.
\newblock In {\em Proceedings of the 31st International Conference on Neural
  Information Processing Systems}, NIPS’17, page 6351–6361, Red Hook, NY,
  USA, 2017. Curran Associates Inc.

\bibitem{pytorchgeomtric}
Matthias Fey and Jan~E. Lenssen.
\newblock Fast graph representation learning with {PyTorch Geometric}.
\newblock In {\em ICLR Workshop on Representation Learning on Graphs and
  Manifolds}, 2019.

\bibitem{gao2018spotlight}
Yuanxiang Gao, Li~Chen, and Baochun Li.
\newblock Spotlight: Optimizing device placement for training deep neural
  networks.
\newblock In {\em International Conference on Machine Learning}, pages
  1676--1684, 2018.

\bibitem{garey1979computers}
Michael~R Garey and David~S Johnson.
\newblock {\em Computers and intractability}, volume 174.
\newblock freeman San Francisco, 1979.

\bibitem{graham1979optimization}
Ronald~L Graham, Eugene~L Lawler, Jan~Karel Lenstra, and AHG~Rinnooy Kan.
\newblock Optimization and approximation in deterministic sequencing and
  scheduling: a survey.
\newblock In {\em Annals of discrete mathematics}, volume~5, pages 287--326.
  Elsevier, 1979.

\bibitem{jeannot2013symbolic}
Emmanuel Jeannot.
\newblock {Symbolic mapping and allocation for the Cholesky factorization on
  NUMA machines: Results and optimizations}.
\newblock {\em The International journal of high performance computing
  applications}, 27(3):283--290, 2013.

\bibitem{GCN}
Thomas~N. Kipf and Max Welling.
\newblock Semi-supervised classification with graph convolutional networks.
\newblock {\em CoRR}, abs/1609.02907, 2016.

\bibitem{multipleResourcemanagement}
Vaibhav Kumar, Siddhant Bhambri, and Prashant~Giridhar Shambharkar.
\newblock Multiple resource management and burst time prediction using deep
  reinforcement learning.
\newblock In {\em Eighth International Conference on Advances in Computing,
  Communication and Information Technology (CCIT)}, page~8, 2019.

\bibitem{leung2004handbook}
Joseph~YT Leung.
\newblock {\em Handbook of scheduling: algorithms, models, and performance
  analysis}.
\newblock CRC press, 2004.

\bibitem{CombinatorialOptimizationwithGraphConvolutionalNetworksandGuidedTreeSearch}
Zhuwen Li, Qifeng Chen, and Vladlen Koltun.
\newblock Combinatorial optimization with graph convolutional networks and
  guided tree search.
\newblock {\em CoRR}, abs/1810.10659, 2018.

\bibitem{resiyrcemanagementwithdeeprl}
Hongzi Mao, Mohammad Alizadeh, Ishai Menache, and Srikanth Kandula.
\newblock Resource management with deep reinforcement learning.
\newblock In {\em Proceedings of the 15th {ACM} Workshop on Hot Topics in
  Networks - {HotNets} '16}, pages 50--56. {ACM} Press, 2016.

\bibitem{mao2016resource}
Hongzi Mao, Mohammad Alizadeh, Ishai Menache, and Srikanth Kandula.
\newblock Resource management with deep reinforcement learning.
\newblock In {\em Proceedings of the 15th ACM Workshop on Hot Topics in
  Networks}, pages 50--56, 2016.

\bibitem{pmlr-v70-mirhoseini17a}
Azalia Mirhoseini, Hieu Pham, Quoc~V. Le, Benoit Steiner, Rasmus Larsen,
  Yuefeng Zhou, Naveen Kumar, Mohammad Norouzi, Samy Bengio, and Jeff Dean.
\newblock Device placement optimization with reinforcement learning.
\newblock In Doina Precup and Yee~Whye Teh, editors, {\em Proceedings of the
  34th International Conference on Machine Learning}, volume~70 of {\em
  Proceedings of Machine Learning Research}, pages 2430--2439, International
  Convention Centre, Sydney, Australia, 06--11 Aug 2017. PMLR.

\bibitem{LearningHeuristicsoverLargeGraphsviaDeepReinforcementLearning}
Akash Mittal, Anuj Dhawan, Sahil Manchanda, Sourav Medya, Sayan Ranu, and
  Ambuj~K. Singh.
\newblock Learning heuristics over large graphs via deep reinforcement
  learning.
\newblock {\em CoRR}, abs/1903.03332, 2019.

\bibitem{A2C}
Volodymyr Mnih, Adri{\`{a}}~Puigdom{\`{e}}nech Badia, Mehdi Mirza, Alex Graves,
  Timothy~P. Lillicrap, Tim Harley, David Silver, and Koray Kavukcuoglu.
\newblock Asynchronous methods for deep reinforcement learning.
\newblock {\em CoRR}, abs/1602.01783, 2016.

\bibitem{paliwal_reinforced_2020}
Aditya Paliwal, Felix Gimeno, Vinod Nair, Yujia Li, Miles Lubin, Pushmeet
  Kohli, and Oriol Vinyals.
\newblock Reinforced genetic algorithm learning for optimizing computation
  graphs, 2020.

\bibitem{schulman2015trust}
John Schulman, Sergey Levine, Pieter Abbeel, Michael Jordan, and Philipp
  Moritz.
\newblock Trust region policy optimization.
\newblock In {\em International conference on machine learning}, pages
  1889--1897, 2015.

\bibitem{sutton1998introduction}
Richard~S Sutton et~al.
\newblock {\em Introduction to reinforcement learning}.
\newblock MIT Press, 1998.

\bibitem{NPhard}
J.D. Ullman.
\newblock {NP-complete scheduling problems}.
\newblock {\em Journal of Computer and System Sciences}, 10(3):384 -- 393,
  1975.

\bibitem{A2Centropy}
Ronald~J. Williams and Jing Peng.
\newblock Function optimization using connectionist reinforcement learning
  algorithms.
\newblock {\em Connection Science}, 3(3):241--268, 1991.

\bibitem{adaptativeDAGtasksscheduling}
Qing Wu, Zhiwei Wu, Yuehui Zhuang, and Yuxia Cheng.
\newblock Adaptive {DAG} tasks scheduling with deep reinforcement learning.
\newblock In Jaideep Vaidya and Jin Li, editors, {\em Algorithms and
  Architectures for Parallel Processing}, volume 11335, pages 477--490.
  Springer International Publishing, 2018.

\bibitem{zd1995}
W.~Zhang and T.G. Dietterich.
\newblock A reinforcement learning approach to job-shop scheduling.
\newblock In {\em Proc. IJCAI}, pages 1114--1120, 1995.

\bibitem{zhou2019gdp}
Yanqi Zhou, Sudip Roy, Amirali Abdolrashidi, Daniel Wong, Peter~C Ma, Qiumin
  Xu, Ming Zhong, Hanxiao Liu, Anna Goldie, Azalia Mirhoseini, et~al.
\newblock Gdp: Generalized device placement for dataflow graphs.
\newblock {\em arXiv preprint arXiv:1910.01578}, 2019.

\end{thebibliography}

\end{document}